\documentclass[10pt,conference]{IEEEtran}
\usepackage{cite}
\usepackage{amsmath,amssymb,amsfonts}
\usepackage{algorithmic}
\usepackage{graphicx}
\usepackage{censor}
\usepackage{subcaption}
\usepackage{textcomp}
\usepackage{xcolor}
\usepackage{booktabs}
\usepackage{multirow}
\usepackage{multicol}
\usepackage{svg}
\usepackage[hidelinks]{hyperref}
\usepackage{url}
\usepackage[acronym]{glossaries}
\usepackage{cleveref}
\usepackage{wrapfig}

\newif\iffinal
\finaltrue

\newacronym{mrz}{MRZ}{Machine-Readable Zone}
\newacronym{ocr}{OCR}{Optical Character Recognition}
\newacronym{pii}{PII}{Personally Identifiable Information}
\newacronym{cnn}{CNN}{Convolutional Neural Network}
\newacronym{iou}{IoU}{Intersection over Union}
\newacronym{mhiou}{mHIoU}{mean Hungarian Intersection over Union}
\newacronym{map}{mAP}{mean Average Precision}

\begin{document}
\title{Anonymization of Documents for Law Enforcement with Machine Learning 
}

\iffinal
\author{\IEEEauthorblockN{1\textsuperscript{st} Manuel Eberhardinger}
\IEEEauthorblockA{
            \textit{Stuttgart Media University}\\
            Stuttgart, Germany \\
            eberhardinger@hdm-stuttgart.de
            }
\and
    \IEEEauthorblockN{1\textsuperscript{st} Patrick Takenaka}
    \IEEEauthorblockA{
            \textit{Stuttgart Media University}\\
            Stuttgart, Germany \\
            takenaka@hdm-stuttgart.de
            }
\and
\IEEEauthorblockN{1\textsuperscript{st} Daniel Grie\ss{}haber}
\IEEEauthorblockA{
            \textit{Stuttgart Media University}\\
            Stuttgart, Germany \\
            griesshaber@hdm-stuttgart.de
            }
\and
\IEEEauthorblockN{2\textsuperscript{nd} Johannes Maucher}
\IEEEauthorblockA{
            \textit{Stuttgart Media University}\\
            Stuttgart, Germany \\
            maucher@hdm-stuttgart.de
            }
}   
\fi

\maketitle

\begin{abstract}

The steadily increasing utilization of data-driven methods and approaches in areas that handle sensitive personal information such as in law enforcement mandates an ever increasing effort in these institutions to comply with data protection guidelines.
In this work, we present a system for automatically anonymizing images of scanned documents, reducing manual effort while ensuring data protection compliance. Our method considers the viability of further forensic processing after anonymization by minimizing automatically redacted areas by combining automatic detection of sensitive regions with knowledge from a manually anonymized reference document. Using a self-supervised image model for instance retrieval of the reference document, our approach requires only one anonymized example to efficiently redact all documents of the same type, significantly reducing processing time.
We show that our approach outperforms both a purely automatic redaction system and also a naive copy-paste scheme of the reference anonymization to other documents on a hand-crafted dataset of ground truth redactions. 

\end{abstract}


\section{Introduction}
With the proliferation of data-driven processing methods into many areas of daily life, data protection has also become increasingly important in recent years. In the European Union, the General Data Protection Regulation~\cite{RegulationEU20162016} was introduced in 2016 to protect personal information. At the same time, similar regulations in the US were updated and made more strict, such as the California Consumer Privacy Act~\cite{CaliforniaConsumerPrivacy2018}. These regulations must be implemented by all companies and government agencies that want to operate in these jurisdictions. For law enforcement authorities compliance is particularly important due to the sensitivity of the data involved, which is commonly achieved by not storing such data, thereby preventing the use of data-driven methods. Still, the utilization of large-scale processing of this data would promise greater efficiency and efficacy in many work processes in this area, but would also require longer and more extensive data storage. To allow storage of this data, one cornerstone is the thorough redaction of all \gls*{pii} in collected documents, e.g. by using named entity recognition~\cite{korytkowski_privacy_2023, biesner_anonymization_2022, juez-hernandez_agora_2023}. Many documents that are relevant to law enforcement---such as travel and identity documents---contain a lot of \gls*{pii} in different forms. The structure and elements on the documents may vary greatly from country to country, and even within one country the layout of documents is updated from time to time, resulting in multiple document models of the same type. Some documents contain barcodes with personal information such as names, pictures and biometric data.  Others have a \gls*{mrz} that also contains the holder's name and date of birth.  These characteristics make purely text-based processing insufficient to achieve data protection compliance. However, for specific models of documents the data and layout are normally well defined. 

In this work, we present a framework for automatic redaction of \gls*{pii} in any document by combining machine learning-based models with image redactions based on the knowledge of domain experts for document anonymization. This results in a system that is capable of large-scale data anonymization given only a single, already anonymized reference document. Given a collection of redacted documents, we are able to obtain the correct reference document using instance retrieval based on image embeddings generated by a self-supervised vision model, therefore not requiring any manual classification of the document beforehand. This minimizes the requirements on the underlying database by not requiring any document metadata for the selection process, making this framework applicable in diverse environments. We show that our joint approach outperforms both a purely data-driven approach and a naive use of the reference redactions as true redactions for the target document by evaluating the framework on a dataset annotated by experts in the specific domain. Our code and data are part of proprietary software and therefore cannot be made available as open source. 

The structure of the paper is as follows: We first give a brief overview of related work on machine learning methods for data privacy in \cref{sec:related-work}. Afterwards, the dataset and the problem domain are presented in \cref{sec:data}. In \cref{sec:method}, we introduce the proposed method for anonymizing documents, followed by a detailed evaluation of the trained models on our dataset in \cref{sec:eval}. Finally, we discuss the limitations and future work of the approach presented in \cref{sec:limit}, and conclude our findings in \cref{sec:conclusion}.

\section{Related Work}
\label{sec:related-work}

Most research for data protection and anonymization focuses on text-based document data~\cite{korytkowski_privacy_2023}. In \cite{biesner_anonymization_2022} several deep learning based methods are studied to anonymize German financial documents. AGORA~\cite{juez-hernandez_agora_2023}, the first anonymization model introduced in a police context in Spain, is based on named entity recognition and used to anonymize Spanish text documents. 

The work that is most similar to ours is~\cite{bouma_document_2020}. However, the main difference is in how the textual content is redacted. Their approach is more rigid and performs \gls*{ocr} to find sensitive, predefined keywords in the document (such as ``Name''), whose position, if found, is used to mask a predefined text region relative to it. In our case, we do not need keyword texts in the document to be able to find regions to be redacted, and we are also more robust against \gls*{ocr} failures, e.g. in the case of handwritten text or varying writing systems. Furthermore, their document type recognition is fixed to five different classes representing different countries, whereas our document recognition is based on instance retrieval and therefore does not need to be retrained to support new document types or models. We also perform a more thorough evaluation based on typical object detection metrics, which provides better insights into the performance of the approach. In follow-up work, Van Rooji et al.~\cite{van_rooij_federated_2022}, presented a federated tool for anonymization and annotation of image data. Similar to our work, they use YOLOv5~\cite{jocher_ultralyticsyolov5_2022}, but evaluate the model only on the Flicker logo dataset~\cite{romberg_scalable_2011} and not on travel and identity documents. Another method for image anonymization was introduced by S{\'a}nchez et al.~\cite{sanchez_automatic_2018} which, however, only used Spanish invoices for evaluation. 

Besides data protection, the field of forensics and law enforcement already utilizes machine learning methods in various other settings, such as for network or mobile forensics, or for detecting malware~\cite{qadir_applications_2021}. Another study used machine learning to analyze sexual predatory online chats~\cite{ngejane_digital_2021}. Other work focuses on person re-identification \cite{ye_deep_2022, wang_spatial-temporal_2019, dietlmeier_how_2021} or identifying the printer for specific documents from a given scan~\cite{lee_printer_2019, guo_printer_2024, takenaka_classification_2024}. 

\section{Data}
\label{sec:data}
\begin{wraptable}[13]{r}{4cm}
    \centering
    \caption{The types and amount of bounding boxes that were annotated in our dataset.}
    \begin{tabular}{l|r}
        \toprule
         \textbf{Type} & \textbf{Number} \\
         \midrule
         Text & $1111$ \\
         Image & $155$ \\
         Signature & $133$ \\
         MRZ & $35$ \\
         Barcode & $18$ \\
    \end{tabular}
    \label{tab:my_label}
\end{wraptable}

Our evaluation data consists of digital scans of six different document types from seven countries, including identity and travel documents, as well as bank transfer forms. 
We manually annotated 206 images of documents containing \gls*{pii}, resulting in 1452 bounding boxes. We anonymized the \gls*{pii} in accordance with the regulations of the European Union, which is defined as ``any information that relates to an identified or identifiable living individual."\footnote{From \url{https://commission.europa.eu/law/law-topic/data-protection/reform/what-personal-data_en}} \Cref{fig:data-samples} on the left shows four redacted samples of this dataset. Each bounding box is labeled with a class that describes its redacted content, which we use to fine-tune our redaction predictions for each type. 


\section{Method}
\label{sec:method}

The redaction process begins by retrieving an already redacted reference document from the database of the same model as the document to be redacted. We find this reference document by measuring the similarity between documents through a trained instance retrieval network. In this work, instance retrieval is used at the category level, since documents matching the country, document type and model are to be returned~\cite{chen_deep_2023}. For this purpose, we learn image representations with deep learning and perform feature matching with the cosine similarity. For its deep feature extraction, we leverage the DinoV2 framework~\cite{oquab_dinov2_2023}, a state-of-the-art self-supervised learning method for images based on vision transformers~\cite{dosovitskiy_image_2020}. Since our domain differs significantly from common public vision datasets, we train a new DinoV2 model from scratch on our own dataset, which is composed of a large collection of various scanned documents. This ensures that the image encoders can extract meaningful features of the document images. 

Next, we use a set of object detection algorithms to predict the locations of text, images of faces, the \gls*{mrz}, and barcodes in the given document. The predicted bounding boxes are then matched, filtered, and adapted to the redactions of the reference document, the details are described in \cref{sec:redaction_prediction}. Based on the class of the redaction (such as text or barcode), we vary our matching process in order to account for their unique characteristics. Since the format of the given and the reference document may not match exactly---i.e. the document may be shifted, scaled, or cropped---we transform the reference redaction bounding boxes using an affine transformation that we estimate by matching key points in both images. For this, we first use the A-KAZE algorithm~\cite{alcantarilla_fast_2013} to find key points in both documents and then match them using a best-fit matcher that minimizes the Hamming distance. Finally, we compute the affine transformation using the RANSAC algorithm~\cite{fischler_random_1981}.

\begin{figure}[tb]
\centering
\includegraphics[width=0.8\linewidth]{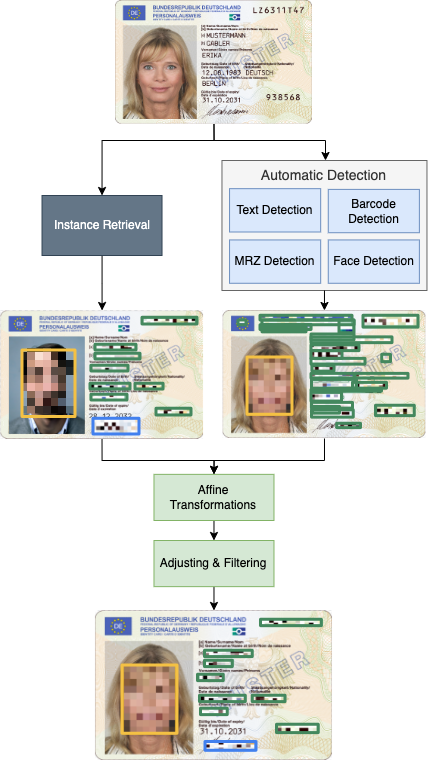}
\caption{The overview of our proposed framework: The input document is used to retrieve already correctly anonymized documents. Various object detection algorithms predict where text, images of faces, the machine-readable zone or barcodes are located on the given document. Afterwards, the predicted bounding boxes are filtered with the reference document and adjusted, e.g. the box width and height. Finally, object types that were not recognized are transferred from the reference document to the input document by using affine transformations. The input document in this figure is a specimen document provided by the German government and does not refer to a real person.}
\label{fig:overview}
\end{figure}

\Cref{fig:overview} shows an overview of the proposed framework, and in the following sections we describe the details of the redaction matching for each type in more detail.

\subsection{Redaction Prediction}\label{sec:redaction_prediction}
\subsubsection{Face Image}
Face detection is a common task not only applicable in forensics and therefore there exist readily available models. We apply a pre-trained YuNet model~\cite{wu_yunet_2023}\footnote{Using \url{https://github.com/opencv/opencv_zoo/tree/main/models/face_detection_yunet}}, which recognizes faces in real-time and is based on a \gls*{cnn}. Since ID documents often contain hard to detect changeable laser images, relying solely on pre-trained face detectors is not sufficient in order to cover all face images in the document. Thus, we also take into account the redactions of face images of the reference document and add any (transformed) reference bounding box that does not overlap with any of the predicted bounding boxes.

\subsubsection{Text}
For detecting text on documents, we use a pre-trained PP OCRv3 model~\cite{li_pp-ocrv3_2022}. PP OCRv3 is a sophisticated and lightweight OCR system based on a collaborative mutual distillation learning framework~\cite{du_pp-ocrv2_2021}, transformers, and \gls*{cnn}. Since our method requires only positional information for the detected text on the document, we only use the differential binarization-based text detection before the rectification and recognition steps of the model.

Unlike to the other types of redactions, for text it is important to distinguish between text that needs to be redacted, and text that is unrelated to \gls*{pii}. We accomplish this by taking into account the transformed text redaction bounding boxes of the reference document. For each reference bounding box, we map it to the predicted bounding box with the highest \gls*{iou}. For any reference bounding box that has an assigned \gls*{iou} greater than $0.1$ we fine-tune its width $w_\mathrm{ref}$ given the best-matching predicted bounding box width $w_\mathrm{pred}$, and its two top-left $x$ coordinates $x_\mathrm{ref}$ and $x_\mathrm{pred}$ as in:

\begin{equation}
    w_\mathrm{ref} = w_\mathrm{pred} - (x_\mathrm{ref} - x_\mathrm{pred})
\end{equation}

This results in bounding boxes that match the reference boxes on the left, while taking the adjusted width from the assigned bounding box prediction. For areas where there was no text prediction, we take the transformed reference bounding box directly, since we have found that some textual descriptions in official documents can be difficult for typical text detectors to detect because the background of the document often contains high-frequency features such as microscript or fine patterns, making forgery in these areas more difficult.

\subsubsection{Barcode}
For barcode detection, we trained a custom model based on the YOLO architecture~\cite{jocher_ultralyticsyolov5_2022} using a custom dataset including different barcode classes, such as linear barcodes, QR barcodes, PDF417 barcodes. Although not needed for the redaction aspect, we have found that combining our barcode detection model with a stamp detection model used for other applications led to better performance in both domains. We match the detected barcodes with the barcodes in the reference document using the same scheme as for the text redactions.

\subsubsection{\gls*{mrz}}
To detect a \gls*{mrz} on the document, we follow the method proposed by Rosebrock~\cite{rosebrockDetectingMachinereadableZones2015} and exploit the fact that the characters are printed in black on a bright background and can therefore be found using morphological closing operations after calculating the Scharr gradient~\cite{scharrOptimalFiltersExtended2007} on the image. We use the PassportEye implementation of this algorithm~\cite{tretyakov_konstantintpassporteye_2024} without the additional step of character recognition. 

We apply the same matching scheme between reference and predicted redactions as for text and barcodes.

\subsubsection{Signature}
Although there exist datasets for training signature detection models~\cite{ultralyticsUltralyticsSignatureDetection} and even pre-trained models~\cite{enginOfflineSignatureVerification2020,st.dev.labStepancoderHandWritenSignatureDetection2024}, our data used in the evaluations contained a large enough domain shift that resulted in these models not showing satisfactory performance. Therefore, the transformed reference bounding boxes are used as final redactions. It is our plan to create a custom dataset in the future to include in the trained YOLO detection model.

\section{Evaluation}
\label{sec:eval}
We first evaluate the instance retrieval component to obtain the reference bounding boxes, then the custom YOLO model for barcode detection, and finally the method to anonymize the documents as a whole.

\subsection{Instance Retrieval}
For the instance retrieval, we evaluated a total of 54 document types from various countries that were provided as data samples. 

DinoV2~\cite{oquab_dinov2_2023} was trained on a large dataset of 320257 scanned documents of different types of documents \iffinal from the State Office of Criminal Investigation in Baden-Württemberg, Germany\fi. In contrast to a normal evaluation of instance retrieval models~\cite{chen_deep_2023}, we only measure whether the instance found with the highest cosine similarity belongs to the same document model as the reference image, since we are only interested in finding one good reference document for the anonymization process.

Across all document models we achieve the highest instance retrieval accuracy of $1.0$, indicating that all documents were correctly assigned.  
Since it has already been shown that DinoV2 is well suited for follow-up tasks without the need to fine-tune the model~\cite{damm2024anomalydinoboostingpatchbasedfewshot, song2024generalpurposeimageencoder}, we consider this to be a plausible result as we train DinoV2 from scratch and visual differences between documents are clearly visible (see \cref{fig:data-samples} for examples).
Therefore, our instance retrieval model can be a suitable basis for obtaining the reference documents for the downstream anonymization process.


\subsection{Barcode Detection}
We evaluated our YOLO model on 4025 images with a total of 7264 bounding boxes for all classes. We report the precision, recall and \gls*{map}. The \gls*{map} refers to the average precision using ten different \gls*{iou} thresholds, which are increased in 10 equal steps from 50 to 95. The \gls*{map} for an \gls*{iou} of 50 is also provided for comparison. The model was trained on 9393 images with a total of 25934 bounding boxes, consisting of 23240 boxes for stamps, 1067 boxes for linear barcodes, 840 boxes for QR barcodes, and 787 boxes for PDF417 barcodes. \Cref{tab:eval-yolo} shows the results. We achieve a \gls*{map} of around $0.90$ for all barcodes, with an even higher \gls*{map}$_{50}$ of $0.97$. We consider this performance to be sufficient for the downstream task of redacting the barcodes.

\begin{table}[t]
\centering
    \caption{The evaluation of the YOLO model for detecting the barcodes and stamps. }
\begin{tabular}{lrrrrr}
\toprule 
        & Instances    &      Precision   &       Recall  &    mAP$_{50}$ &  mAP \\
                 Class    &               &          &          &          &      \\ 
\midrule
            all    &       7264    &  0.968   &   0.947  &     0.97 &     0.872   \\
           Stamp    &       6452    &  0.933   &   0.871  &    0.937 &     0.769 \\
    Linear Barcode    &       341     &  0.973   &   0.977  &     0.98 &      0.88 \\
      QR Barcode    &       195     &  0.969   &   0.954  &    0.969 &     0.909 \\
     PDF417 Barcode    &       276     &  0.996   &   0.986  &    0.992 &     0.932 \\
\bottomrule
\end{tabular}
    \label{tab:eval-yolo}
\end{table}

\subsection{Evaluation of the Document Anonymizations}
\begin{figure*}[t]
    \includegraphics[width=\linewidth]{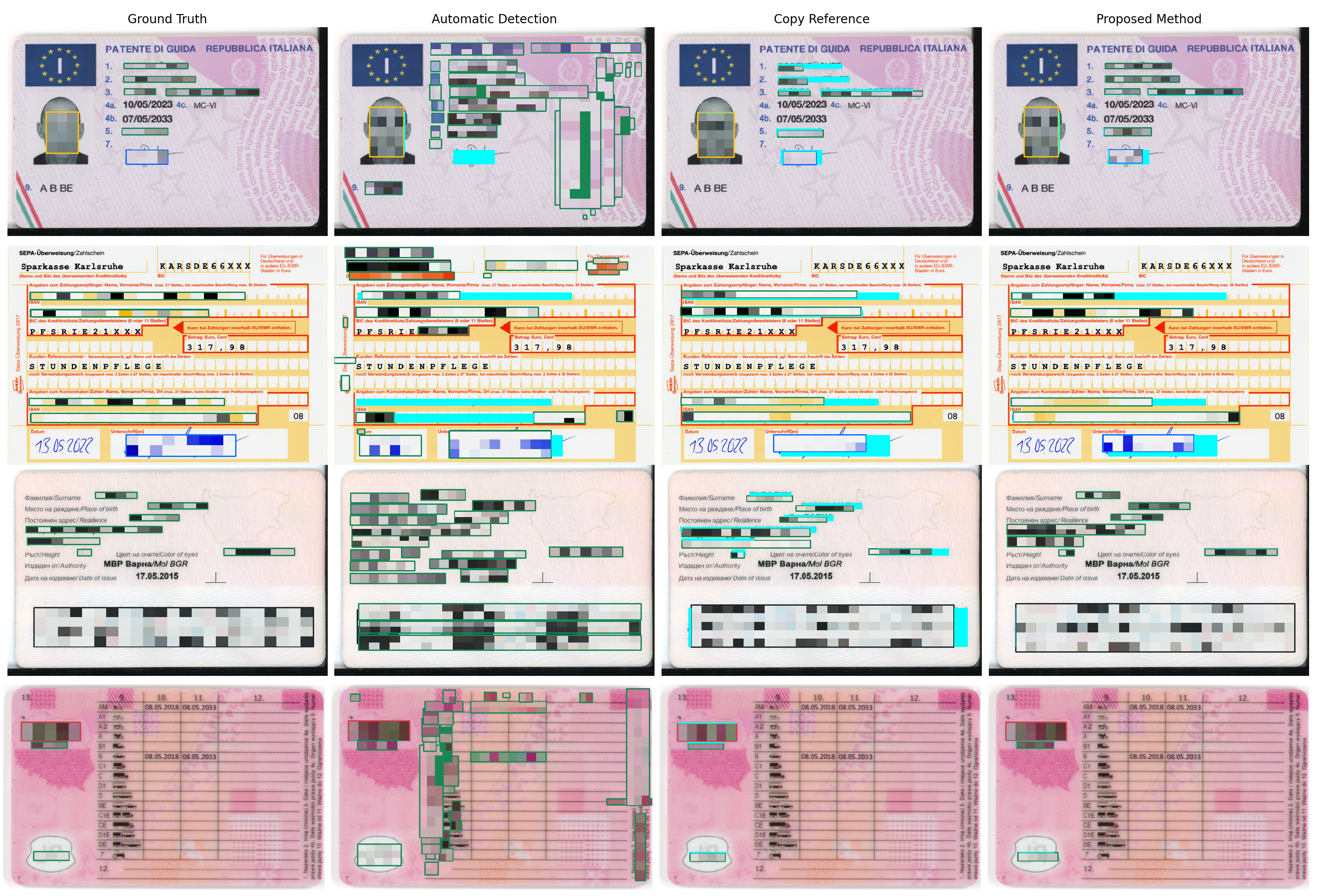}
  \caption{Four images of different document types and the anonymization of the different methods. On the left, we show the ground truth, followed by the two baseline methods and our proposed method. The color of the redacted bounding box indicates the different elements recognized. Yellow boxes are faces, green indicates text, blue for signatures, black for the \gls*{mrz} and red for barcodes. The turquoise color shows the difference between the predicted bounding box and the ground truth. From top to bottom, we show the front of an Italian driver's license, a German bank transfer form, the front of a Bulgarian ID card and the back of a Polish driver's license.}
  \label{fig:data-samples}
\end{figure*}
To evaluate our proposed anonymization method, we use the dataset described in \cref{sec:data}.

We use two common metrics for object detection: (1) \gls*{mhiou}, which finds the best match between predicted and ground truth boxes as measured by \gls*{iou}, and (2) \gls*{map}. While \gls*{map} provides insight into the overall detection performance, the \gls*{mhiou} allows us to evaluate how well our method aligns the redacted areas with the ground truth bounding boxes of the \gls*{pii} in the documents. We report two baselines: The first takes the automatically predicted bounding boxes as final predictions without considering the reference document, the second copies the reference bounding boxes directly to the given document to be redacted. Since our dataset contains ground truths for all annotated documents, we have multiple candidates for the reference document. To ablate any effects that a particular document may have on the results, we randomly select one document from all candidates as a reference. The two baselines represent the two components from which our joint method consists of and each by itself would be a reasonable but naive solution to the anonymization problem. However, since the use of only automatically detected features does not rely on a pre-annotated reference document, this baseline method can also be used to bootstrap the database of reference documents, requiring only adjustments by a data protection expert.

\Cref{fig:data-samples} shows four redacted document samples from the dataset and the corresponding redactions predicted by the different methods. On the left the ground truth is shown, followed by the two baselines and our proposed method. The turquoise color shows the difference of the predicted bounding box to the ground truth. For the back of the Polish driver's license and the front of a Bulgarian ID card, our proposed method is able to anonymize all \gls*{pii}, which shows the improvements over the baselines. 

\begin{table}[t]
    \centering
    \caption{The results of the proposed method averaged over all document types. We show aggregated and individual scores for the different redaction classes.}
\begin{tabular}{lrrrr}
\toprule
\textbf{Method} &  \textbf{mHIoU} & \textbf{mAP} & \textbf{mAP$_{50}$} & \textbf{mAP$_{75}$}  \\
\midrule
Automatic Detection & 0.486 & 0.166 & 0.319 & 0.140 \\
Copy Reference & 0.571 & 0.245 & 0.520 & 0.195 \\
Proposed Method & \textbf{0.741} & \textbf{0.445} & \textbf{0.721} & \textbf{0.423} \\
\bottomrule
\end{tabular}
    \label{tab:eval-mean-overall}
\end{table}

\begin{table}[t]
    \centering
    \caption{The results of the proposed method individually for each redaction class.}
\begin{tabular}{lrrrrr}
\toprule
 & \multicolumn{5}{c}{\textbf{mAP}} \\
\textbf{Method} &  Image & Text & Barcode & MRZ & Signature  \\
\midrule
Automatic Detection & 0.346 & 0.019 & 0.763 & 0.576 & N/A \\
Copy Reference &  0.319 & 0.153 & 0.515 & 0.781 & 0.054 \\
Proposed Method & \textbf{0.403} & \textbf{0.527} & \textbf{0.879} & \textbf{0.941} & \textbf{0.087} \\
\bottomrule
\end{tabular}
    \label{tab:eval-mean-individual}
\end{table}

\Cref{tab:eval-mean-overall,tab:eval-mean-individual} show the final metrics for the entire dataset. Overall, our proposed method outperforms the two baselines by a large margin for all metrics except for the signature class. For signatures the margin is smaller, as no detection model is currently used.  Nevertheless, it clearly shows the positive impact of the affine transformation based on key point matching between the reference document and the given document, as opposed to simply copying the reference redactions. 

\begin{table*}[t]
    \centering
    \caption{The detailed results of the proposed method for different document types. Some types do not have all of the redaction classes present and their scores are therefore not available.}
\begin{tabular}{p{2cm}lrrrr|rrrrr}
\toprule
& & & & & & \multicolumn{5}{c}{\textbf{mAP}} \\
\textbf{Document Type} & \textbf{Method} &  \textbf{mHIoU} & \textbf{mAP} & \textbf{mAP$_{50}$} & \textbf{mAP$_{75}$} &  Image & Text & Barcode & MRZ & Signature  \\
\midrule
\multirow{3}{2cm}{Belgian Drivers License Front} & Automatic Detection & 0.373 & 0.097 & 0.198 & 0.055 & 0.289 & 0.003 & N/A & N/A & 0.000 \\
 & Copy Reference & 0.579 & 0.207 & 0.478 & 0.158 & 0.397 & 0.197 & N/A & N/A & \textbf{0.028} \\
 & Proposed Method & \textbf{0.702} & \textbf{0.360} & \textbf{0.621} & \textbf{0.333} & \textbf{0.482} & \textbf{0.572} & N/A & N/A & 0.026 \\
\midrule
\multirow{3}{2cm}{Polish Drivers License Front} & Automatic Detection & 0.429 & 0.095 & 0.260 & 0.014 & 0.279 & 0.005 & N/A & N/A & 0.000 \\
& Copy Reference & 0.593 & 0.159 & 0.411 & 0.122 & 0.237 & 0.200 & N/A & N/A & \textbf{0.039} \\
& Proposed Method & \textbf{0.751} & \textbf{0.306} & \textbf{0.564} & \textbf{0.280} & \textbf{0.289} & \textbf{0.592} & N/A & N/A & 0.037 \\
\midrule
\multirow{3}{2cm}{Polish Drivers License Back} & Automatic Detection & 0.614 & 0.382 & 0.504 & 0.457 & N/A & 0.001 & 0.763 & N/A & N/A \\
& Copy Reference & 0.652 & 0.336 & 0.773 & 0.229 & N/A & 0.158 & 0.515 & N/A & N/A \\
& Proposed Method & \textbf{0.864} & \textbf{0.764} & \textbf{1.000} & \textbf{0.929} & N/A & \textbf{0.650} & \textbf{0.879} & N/A & N/A \\
\midrule
\multirow{3}{2cm}{Spanish Drivers License Front} & Automatic Detection & 0.472 & 0.126 & 0.375 & 0.006 & \textbf{0.349} & 0.028 & N/A & N/A & 0.000 \\
& Copy Reference & 0.499 & 0.051 & 0.163 & 0.022 & 0.026 & 0.103 & N/A & N/A & \textbf{0.023} \\
& Proposed Method & \textbf{0.776} & \textbf{0.358} & \textbf{0.674} & \textbf{0.307} & \textbf{0.349} & \textbf{0.714} & N/A & N/A & 0.011 \\
\midrule
\multirow{3}{2cm}{Polish Residence Permit Front} & Automatic Detection & 0.515 & 0.107 & 0.361 & 0.008 & 0.293 & 0.029 & N/A & N/A & 0.000 \\
& Copy Reference & 0.550 & 0.176 & 0.471 & 0.048 & \textbf{0.369} & 0.119 & N/A & N/A & \textbf{0.042} \\
& Proposed Method & \textbf{0.678} & \textbf{0.262} & \textbf{0.622} & \textbf{0.215} & 0.293 & \textbf{0.468} & N/A & N/A & 0.026 \\
\midrule
\multirow{3}{2cm}{Polish Residence Permit Back} & Automatic Detection & 0.492 & 0.142 & 0.223 & 0.178 & N/A & 0.012 & N/A & 0.272 & N/A \\
& Copy Reference & 0.572 & 0.482 & 0.684 & 0.518 & N/A & 0.107 & N/A & 0.857 & N/A \\
& Proposed Method & \textbf{0.755} & \textbf{0.711} & \textbf{0.908} & \textbf{0.694} & N/A & \textbf{0.447} & N/A & \textbf{0.975} & N/A \\
\midrule
\multirow{3}{2cm}{Bulgarian ID Card Front} & Automatic Detection & 0.443 & 0.098 & 0.260 & 0.004 & 0.277 & 0.019 & N/A & N/A & 0.000 \\
& Copy Reference & 0.541 & 0.188 & 0.465 & 0.100 & 0.438 & 0.081 & N/A & N/A & 0.045 \\
& Proposed Method & \textbf{0.790} & \textbf{0.389} & \textbf{0.783} & \textbf{0.245} & \textbf{0.441} & \textbf{0.607} & N/A & N/A & \textbf{0.120} \\
\midrule
\multirow{3}{2cm}{Bulgarian ID Card Back} & Automatic Detection & 0.556 & 0.362 & 0.563 & 0.500 & N/A & 0.021 & N/A & 0.703 & N/A \\
& Copy Reference & 0.592 & 0.389 & 0.683 & 0.384 & N/A & 0.115 & N/A & 0.664 & N/A \\
& Proposed Method & \textbf{0.760} & \textbf{0.680} & \textbf{0.943} & \textbf{0.707} & N/A & \textbf{0.464} & N/A & \textbf{0.896} & N/A \\
\midrule
\multirow{3}{2cm}{German Bank Transfer Form Front} & Automatic Detection & 0.343 & 0.003 & 0.011 & 0.001 & N/A & 0.006 & N/A & N/A & 0.000 \\
& Copy Reference & \textbf{0.673} & \textbf{0.201} & \textbf{0.639} & \textbf{0.084} & N/A & \textbf{0.281} & N/A & N/A & 0.121 \\
& Proposed Method & 0.569 & 0.170 & 0.462 & 0.076 & N/A & 0.096 & N/A & N/A & \textbf{0.244} \\
\midrule
\multirow{3}{2cm}{Slovenian Residence Permit Front} & Automatic Detection & 0.625 & 0.164 & 0.390 & 0.007 & 0.450 & 0.041 & N/A & N/A & 0.000 \\
& Copy Reference & 0.657 & 0.285 & 0.730 & 0.197 & \textbf{0.503} & 0.220 & N/A & N/A & 0.132 \\
& Proposed Method & \textbf{0.796} & \textbf{0.465} & \textbf{0.879} & \textbf{0.354} & 0.450 & \textbf{0.720} & N/A & N/A & \textbf{0.224} \\
\midrule
\multirow{3}{2cm}{Slovenian Residence Permit Back} & Automatic Detection & 0.677 & 0.417 & 0.672 & 0.501 & N/A & 0.080 & N/A & 0.754 & N/A \\
& Copy Reference & 0.674 & 0.563 & 0.798 & 0.613 & N/A & 0.304 & N/A & 0.822 & N/A \\
& Proposed Method & \textbf{0.882} & \textbf{0.863} & \textbf{0.980} & \textbf{0.980} & N/A & \textbf{0.774} & N/A & \textbf{0.951} & N/A \\
\midrule
\multirow{3}{2cm}{Italian Drivers License Front} & Automatic Detection & 0.500 & 0.163 & 0.332 & 0.084 & 0.482 & 0.007 & N/A & N/A & 0.000 \\
& Copy Reference & 0.465 & 0.116 & 0.386 & 0.048 & 0.265 & 0.081 & N/A & N/A & 0.003 \\
& Proposed Method & \textbf{0.771} & \textbf{0.402} & \textbf{0.673} & \textbf{0.377} & \textbf{0.515} & \textbf{0.687} & N/A & N/A & \textbf{0.004} \\
\midrule
\multirow{3}{2cm}{Italian Drivers License Back} & Automatic Detection & 0.285 & 0.001 & 0.003 & 0.000 & N/A & 0.001 & N/A & N/A & N/A \\
& Copy Reference & 0.376 & 0.025 & 0.080 & \textbf{0.008} & N/A & 0.025 & N/A & N/A & N/A \\
& Proposed Method & \textbf{0.535} & \textbf{0.059} & \textbf{0.267} & \textbf{0.008} & N/A & \textbf{0.059} & N/A & N/A & N/A \\
\bottomrule
\end{tabular}
    \label{tab:eval-full}
\end{table*}

\Cref{tab:eval-full} shows the detailed evaluation for specific document models and each redaction class. In almost all cases, our method is better than the baselines. The improvements for the text class are particularly visible. This is easily explained by the structure of the documents, since all evaluated documents contain a lot of text that is not related to personal information and therefore does not need to be anonymized. The automatic detection of the bounding boxes does not have any information about which text needs to be anonymized and which does not. Therefore, the entire text is anonymized. On the other hand, the ground truth baseline only transfers the bounding boxes to other documents without taking into account the length of the text in the target document. For example, it is rarely the case that the length of a person's name or city matches between documents. In addition, affine transformations help with differences in the positioning of document elements, such as when the image is scanned upside down or slightly rotated.

The only document model where the ground truth baseline performs slightly better is the bank transfer form. For this document type, text recognition did not work well, as many names were not recognized correctly and were truncated in the middle. We hypothesize that the font and glyph spacing of the text was not sufficiently present in the training dataset of the OCR model~\cite{li_pp-ocrv3_2022}.

\section{Limitations \& Future Work}
\label{sec:limit}
Despite considerable improvements, there are still some limitations that need to be addressed. Currently, our system mainly uses pre-trained models that are not fine-tuned on our data to match the use case. Training custom models for each anonymization type would mitigate the issues that were visible with the signature element type and also with the OCR on the bank transfer form. In particular, our current approach would benefit greatly from a working signature detection model, as using only the transformed reference document redactions did not yield sufficient performance.

Finally, even with our approach, it is still recommended that a human verifies the correctness of the predicted redactions, as data privacy is a sensitive issue and false negative redactions would pose a breach of its compliance.

\section{Conclusion}
\label{sec:conclusion}
We presented a framework for anonymizing personal data on various scanned documents. We have combined a set of automatic document element detectors with redactions from a reference document to produce anonymizations that are better than both; solely relying on the automatic detectors alone, as well as simply applying the reference redactions directly. We believe this is a major step forward in reducing the time required to process sensitive and confidential images containing personal information. In addition, the presented framework is useful not only for government agencies, but also for private companies, as data regulations are becoming more stringent worldwide.

\section*{Ethical Impact Statement}
This work directly addresses a key issue of ethical AI systems: Data protection. Our main intention of this work is to improve upon current automatic anonymization techniques in order to (1) prevent information leakage in large-scale data, and (2) to allow training of AI models without capturing biases of the personal information in the data. 
In particular, the latter point is a major concern in many domains where decision makers---for example, in law enforcement---are increasingly relying on trained models to take advantage of statistical relationships observed in the respective domain.

\iffinal
\section*{Acknowledgment}
We would like to thank the Document Forensics Department of the Baden Württemberg State Office of Criminal Investigations for their assistance in providing the scanned documents that were necessary for this project. Special recognition is extended to the head of the department, Rolf Fauser, whose personal commitment and insightful expertise have greatly facilitated our work.
We would also like to thank Janina Mattes and Kien Dang for creating the initial dataset for training the YOLO model.
\fi

\bibliographystyle{./IEEEtran}
\bibliography{./IEEEabrv,./main.bib}

\begin{thebibliography}{10}
\providecommand{\url}[1]{#1}
\csname url@samestyle\endcsname
\providecommand{\newblock}{\relax}
\providecommand{\bibinfo}[2]{#2}
\providecommand{\BIBentrySTDinterwordspacing}{\spaceskip=0pt\relax}
\providecommand{\BIBentryALTinterwordstretchfactor}{4}
\providecommand{\BIBentryALTinterwordspacing}{\spaceskip=\fontdimen2\font plus
\BIBentryALTinterwordstretchfactor\fontdimen3\font minus \fontdimen4\font\relax}
\providecommand{\BIBforeignlanguage}[2]{{%
\expandafter\ifx\csname l@#1\endcsname\relax
\typeout{** WARNING: IEEEtran.bst: No hyphenation pattern has been}%
\typeout{** loaded for the language `#1'. Using the pattern for}%
\typeout{** the default language instead.}%
\else
\language=\csname l@#1\endcsname
\fi
#2}}
\providecommand{\BIBdecl}{\relax}
\BIBdecl

\bibitem{RegulationEU20162016}
\BIBentryALTinterwordspacing
``Regulation ({{EU}}) 2016/679 of the {{European Parliament}} and of the {{Council}} of 27 {{April}} 2016 on the protection of natural persons with regard to the processing of personal data and on the free movement of such data, and repealing {{Directive}} 95/46/{{EC}} ({{General Data Protection Regulation}}) ({{Text}} with {{EEA}} relevance).'' [Online]. Available: \url{http://data.europa.eu/eli/reg/2016/679/oj/eng}
\BIBentrySTDinterwordspacing

\bibitem{CaliforniaConsumerPrivacy2018}
\BIBentryALTinterwordspacing
``California {{Consumer Privacy Act}} of 2018.'' [Online]. Available: \url{https://leginfo.legislature.ca.gov/faces/codes_displayText.xhtml?division=3.&part=4.&lawCode=CIV&title=1.81.5}
\BIBentrySTDinterwordspacing

\bibitem{korytkowski_privacy_2023}
M.~Korytkowski, J.~Nowak, R.~Scherer, and W.~Wei, ``Privacy {{Preserving}} by~{{Removing Sensitive Data}} from~{{Documents}} with~{{Fully Convolutional Networks}},'' in \emph{Artificial {{Intelligence}} and {{Soft Computing}}}, L.~Rutkowski, R.~Scherer, M.~Korytkowski, W.~Pedrycz, R.~Tadeusiewicz, and J.~M. Zurada, Eds.\hskip 1em plus 0.5em minus 0.4em\relax Cham: Springer International Publishing, 2023, pp. 277--285.

\bibitem{biesner_anonymization_2022}
D.~Biesner, R.~Ramamurthy, R.~Stenzel, M.~L{\"u}bbering, L.~Hillebrand, A.~Ladi, M.~Pielka, R.~Loitz, C.~Bauckhage, and R.~Sifa, ``Anonymization of {{German}} financial documents using neural network-based language models with contextual word representations,'' \emph{International Journal of Data Science and Analytics}, vol.~13, no.~2, pp. 151--161, Mar. 2022.

\bibitem{juez-hernandez_agora_2023}
R.~{Juez-Hernandez}, L.~{Quijano-S{\'a}nchez}, F.~Liberatore, and J.~G{\'o}mez, ``{{AGORA}}: {{An}} intelligent system for the anonymization, information extraction and automatic mapping of sensitive documents,'' \emph{Applied Soft Computing}, vol. 145, p. 110540, Sep. 2023.

\bibitem{bouma_document_2020}
H.~Bouma, R.~Pruim, A.~V. Rooijen, J.-M. ten Hove, J.~van Mil, and B.~Kromhout, ``Document anonymization for border guards and immigration services,'' in \emph{Counterterrorism, {{Crime Fighting}}, {{Forensics}}, and {{Surveillance Technologies IV}}}, vol. 11542.\hskip 1em plus 0.5em minus 0.4em\relax SPIE, Sep. 2020, pp. 68--75.

\bibitem{van_rooij_federated_2022}
S.~Van~Rooij, H.~Bouma, J.~Van~Mil, and J.-M. Ten~Hove, ``Federated tool for anonymization and annotation in image data,'' in \emph{Counterterrorism, {{Crime Fighting}}, {{Forensics}}, and {{Surveillance Technologies VI}}}, H.~Bouma, R.~J. Stokes, Y.~Yitzhaky, and R.~Prabhu, Eds.\hskip 1em plus 0.5em minus 0.4em\relax Berlin, Germany: SPIE, Oct. 2022, p.~13.

\bibitem{jocher_ultralyticsyolov5_2022}
G.~Jocher, {Ayush Chaurasia}, A.~Stoken, J.~Borovec, {NanoCode012}, {Yonghye Kwon}, {Kalen Michael}, {TaoXie}, {Jiacong Fang}, {Imyhxy}, Lorna, {Zeng Yifu}, C.~Wong, {Abhiram V}, D.~Montes, {Zhiqiang Wang}, C.~Fati, {Jebastin Nadar}, {Laughing}, {UnglvKitDe}, V.~Sonck, {Tkianai}, {YxNONG}, P.~Skalski, A.~Hogan, {Dhruv Nair}, M.~Strobel, and M.~Jain, ``Ultralytics/yolov5: V7.0 - {{YOLOv5 SOTA Realtime Instance Segmentation}},'' Zenodo, Nov. 2022.

\bibitem{romberg_scalable_2011}
S.~Romberg, L.~G. Pueyo, R.~Lienhart, and R.~Van~Zwol, ``Scalable logo recognition in real-world images,'' in \emph{Proceedings of the 1st {{ACM International Conference}} on {{Multimedia Retrieval}}}.\hskip 1em plus 0.5em minus 0.4em\relax Trento Italy: ACM, Apr. 2011, pp. 1--8.

\bibitem{sanchez_automatic_2018}
{\'A}.~S{\'a}nchez, J.~F. V{\'e}lez, J.~S{\'a}nchez, and A.~B. Moreno, ``Automatic {{Anonymization}} of {{Printed-Text Document Images}},'' in \emph{Image and {{Signal Processing}}}, A.~Mansouri, A.~El~Moataz, F.~Nouboud, and D.~Mammass, Eds.\hskip 1em plus 0.5em minus 0.4em\relax Cham: Springer International Publishing, 2018, pp. 145--152.

\bibitem{qadir_applications_2021}
S.~Qadir and B.~Noor, ``Applications of {{Machine Learning}} in {{Digital Forensics}},'' in \emph{2021 {{International Conference}} on {{Digital Futures}} and {{Transformative Technologies}} ({{ICoDT2}})}, May 2021, pp. 1--8.

\bibitem{ngejane_digital_2021}
C.~H. Ngejane, J.~H.~P. Eloff, T.~J. Sefara, and V.~N. Marivate, ``Digital forensics supported by machine learning for the detection of online sexual predatory chats,'' \emph{Forensic Science International: Digital Investigation}, vol.~36, p. 301109, Mar. 2021.

\bibitem{ye_deep_2022}
M.~Ye, J.~Shen, G.~Lin, T.~Xiang, L.~Shao, and S.~C.~H. Hoi, ``Deep {{Learning}} for {{Person Re-Identification}}: {{A Survey}} and {{Outlook}},'' \emph{IEEE Transactions on Pattern Analysis and Machine Intelligence}, vol.~44, no.~6, pp. 2872--2893, Jun. 2022.

\bibitem{wang_spatial-temporal_2019}
G.~Wang, J.~Lai, P.~Huang, and X.~Xie, ``Spatial-{{Temporal Person Re-Identification}},'' \emph{Proceedings of the AAAI Conference on Artificial Intelligence}, vol.~33, no.~01, pp. 8933--8940, Jul. 2019.

\bibitem{dietlmeier_how_2021}
J.~Dietlmeier, J.~Antony, K.~McGuinness, and N.~E. O'Connor, ``How important are faces for person re-identification?'' in \emph{2020 25th {{International Conference}} on {{Pattern Recognition}} ({{ICPR}})}, Jan. 2021, pp. 6912--6919.

\bibitem{lee_printer_2019}
S.-H. Lee and H.-Y. Lee, ``{Printer Identification Methods Using Global and Local Feature-Based Deep Learning},'' \emph{KIPS Transactions on Software and Data Engineering}, vol.~8, no.~1, pp. 37--44, 2019.

\bibitem{guo_printer_2024}
Z.~Guo, S.~Wang, Z.~Zheng, and K.~Sun, ``Printer source identification of quick response codes using residual attention network and smartphones,'' \emph{Engineering Applications of Artificial Intelligence}, vol. 131, p. 107822, May 2024.

\bibitem{takenaka_classification_2024}
P.~Takenaka, M.~Eberhardinger, D.~Grie{\ss}haber, and J.~Maucher, ``Classification of {{Inkjet Printers}} based on {{Droplet Statistics}},'' in \emph{2024 {{International Joint Conference}} on {{Neural Networks}} ({{IJCNN}})}, Jun. 2024, pp. 1--7.

\bibitem{chen_deep_2023}
W.~Chen, Y.~Liu, W.~Wang, E.~M. Bakker, T.~Georgiou, P.~Fieguth, L.~Liu, and M.~S. Lew, ``Deep {{Learning}} for {{Instance Retrieval}}: {{A Survey}},'' \emph{IEEE Transactions on Pattern Analysis and Machine Intelligence}, vol.~45, no.~6, pp. 7270--7292, Jun. 2023.

\bibitem{oquab_dinov2_2023}
M.~Oquab, T.~Darcet, T.~Moutakanni, H.~V. Vo, M.~Szafraniec, V.~Khalidov, P.~Fernandez, D.~Haziza, F.~Massa, A.~{El-Nouby}, M.~Assran, N.~Ballas, W.~Galuba, R.~Howes, P.-Y. Huang, S.-W. Li, I.~Misra, M.~Rabbat, V.~Sharma, G.~Synnaeve, H.~Xu, H.~Jegou, J.~Mairal, P.~Labatut, A.~Joulin, and P.~Bojanowski, ``{{DINOv2}}: {{Learning Robust Visual Features}} without {{Supervision}},'' \emph{Transactions on Machine Learning Research}, Jul. 2023.

\bibitem{dosovitskiy_image_2020}
A.~Dosovitskiy, L.~Beyer, A.~Kolesnikov, D.~Weissenborn, X.~Zhai, T.~Unterthiner, M.~Dehghani, M.~Minderer, G.~Heigold, S.~Gelly, J.~Uszkoreit, and N.~Houlsby, ``An {{Image}} is {{Worth}} 16x16 {{Words}}: {{Transformers}} for {{Image Recognition}} at {{Scale}},'' in \emph{International {{Conference}} on {{Learning Representations}}}, Oct. 2020.

\bibitem{alcantarilla_fast_2013}
P.~Alcantarilla, J.~Nuevo, and A.~Bartoli, ``Fast {{Explicit Diffusion}} for {{Accelerated Features}} in {{Nonlinear Scale Spaces}},'' in \emph{Procedings of the {{British Machine Vision Conference}} 2013}.\hskip 1em plus 0.5em minus 0.4em\relax Bristol: British Machine Vision Association, 2013, pp. 13.1--13.11.

\bibitem{fischler_random_1981}
M.~A. Fischler and R.~C. Bolles, ``Random sample consensus: A paradigm for model fitting with applications to image analysis and automated cartography,'' \emph{Commun. ACM}, vol.~24, no.~6, pp. 381--395, Jun. 1981.

\bibitem{wu_yunet_2023}
W.~Wu, H.~Peng, and S.~Yu, ``{{YuNet}}: {{A Tiny Millisecond-level Face Detector}},'' \emph{Machine Intelligence Research}, vol.~20, no.~5, pp. 656--665, Oct. 2023.

\bibitem{li_pp-ocrv3_2022}
C.~Li, W.~Liu, R.~Guo, X.~Yin, K.~Jiang, Y.~Du, Y.~Du, L.~Zhu, B.~Lai, X.~Hu, D.~Yu, and Y.~Ma, ``{{PP-OCRv3}}: {{More Attempts}} for the {{Improvement}} of {{Ultra Lightweight OCR System}},'' 2022.

\bibitem{du_pp-ocrv2_2021}
Y.~Du, C.~Li, R.~Guo, C.~Cui, W.~Liu, J.~Zhou, B.~Lu, Y.~Yang, Q.~Liu, X.~Hu, D.~Yu, and Y.~Ma, ``{{PP-OCRv2}}: {{Bag}} of {{Tricks}} for {{Ultra Lightweight OCR System}},'' Oct. 2021.

\bibitem{rosebrockDetectingMachinereadableZones2015}
A.~Rosebrock, ``Detecting machine-readable zones in passport images,'' Nov. 2015.

\bibitem{scharrOptimalFiltersExtended2007}
H.~Scharr, ``Optimal {{Filters}} for {{Extended Optical Flow}},'' in \emph{Complex {{Motion}}}, B.~J{\"a}hne, R.~Mester, E.~Barth, and H.~Scharr, Eds.\hskip 1em plus 0.5em minus 0.4em\relax Berlin, Heidelberg: Springer, 2007, pp. 14--29.

\bibitem{tretyakov_konstantintpassporteye_2024}
K.~Tretyakov, ``Konstantint/{{PassportEye}},'' Aug. 2024.

\bibitem{ultralyticsUltralyticsSignatureDetection}
Ultralytics, ``Ultralytics {{Signature Detection Dataset}},'' https://docs.ultralytics.com/datasets/detect/signature.

\bibitem{enginOfflineSignatureVerification2020}
D.~Engin, A.~Kantarci, S.~Arslan, and H.~K. Ekenel, ``Offline {{Signature Verification}} on {{Real-World Documents}},'' in \emph{Proceedings of the {{IEEE}}/{{CVF Conference}} on {{Computer Vision}} and {{Pattern Recognition Workshops}}}, 2020, pp. 808--809.

\bibitem{st.dev.labStepancoderHandWritenSignatureDetection2024}
St.Dev.Lab, ``Stepan-coder/{{HandWritenSignatureDetection}},'' Sep. 2024.

\bibitem{damm2024anomalydinoboostingpatchbasedfewshot}
\BIBentryALTinterwordspacing
S.~Damm, M.~Laszkiewicz, J.~Lederer, and A.~Fischer, ``Anomalydino: Boosting patch-based few-shot anomaly detection with dinov2,'' 2024. [Online]. Available: \url{https://arxiv.org/abs/2405.14529}
\BIBentrySTDinterwordspacing

\bibitem{song2024generalpurposeimageencoder}
\BIBentryALTinterwordspacing
X.~Song, X.~Xu, and P.~Yan, ``General purpose image encoder dinov2 for medical image registration,'' 2024. [Online]. Available: \url{https://arxiv.org/abs/2402.15687}
\BIBentrySTDinterwordspacing

\end{thebibliography}

\end{document}